\documentclass[conference]{IEEEtran}
\IEEEoverridecommandlockouts
\usepackage{cite}
\usepackage{amsmath,amssymb,amsfonts}
\usepackage{algorithmic}
\usepackage{graphicx}
\usepackage{textcomp}
\usepackage{xcolor}
\def\BibTeX{{\rm B\kern-.05em{\sc i\kern-.025em b}\kern-.08em
    T\kern-.1667em\lower.7ex\hbox{E}\kern-.125emX}}
\begin{document}

\title{Data Processing Techniques for Modern Multimodal Models}

\author{
\IEEEauthorblockN{Yinheng Li\textsuperscript{1}}
\IEEEauthorblockA{
\textit{Columbia University}\\
New York, NY, USA \\
yl4039@columbia.edu}
\and
\IEEEauthorblockN{Han Ding\textsuperscript{1}}
\IEEEauthorblockA{
\textit{Columbia University}\\
New York, NY, USA \\
hd2412@columbia.edu}
\and
\IEEEauthorblockN{Hang Chen\textsuperscript}
\IEEEauthorblockA{
\textit{New York University}\\
New York, NY, USA \\
hc2798@nyu.edu}
}

\maketitle

\footnotetext[1]{These authors contributed equally to this work. The order is completely random.}

\begin{abstract}
Data processing plays an significant role in current multimodal model training. In this paper. we provide an comprehensive review of common data processing techniques used in modern multimodal model training with a focus on diffusion models and multimodal large language models (MLLMs). We summarized all techniques into four categories: data quality, data quantity, data distribution and data safety. We further present our findings in the choice of data process methods in different type of models. This study aims to provide guidance to  multimodal models developers with effective data processing techniques.

\end{abstract}


\section{Introduction}
The recent success of multimodal models heavily relies on the use of large-scale datasets such as LAION-5B \cite{schuhmann2022LAION}, Conceptual Caption-3M \cite{sharma-etal-2018-conceptual}. In addition to annotated and curated datasets, web-crawled data has increasingly played a significant role in training datasets. The first step of model training is usually data collection and data cleaning. Having a high quality dataset has proven to be essential for developing robust and high-performing models \cite{budach2022effects}. 

In this work, we surveyed a list of influential multimodal model research at the time of this paper. We review and categorize the commonly used data processing methods and summarize a standard data processing framework for training multimodal models. This framework addresses topics such as data quality, data distribution, and data safety. We believe this work will provide valuable guidance and insights for future efforts to train or fine-tune multimodal models.

\section{Background}
\subsubsection{Multimodal Generative models} 
Multimodal models have seen significant development in recent years. While multimodal models often refer to those that incorporate more than one modality, vision-language models are the most extensively studied areas \cite{du2022survey, duan2022multimodal}. In this work, we focus on data processing techniques for vision-language models, and some of these concepts can be also applied to models involving other modalities.

There are two major categories of vision-language models: discriminative models and generative models. Discriminative models typically use early fusion or late fusion architectures to learn meaningful representations for downstream tasks such as classification and ranking. Most early work centered on discriminative models, including ALBEF \cite{li2021align}, ALIGN \cite{jia2021scaling}, and CLIP \cite{radford2021learning}.

Generative models, on the other hand, aim to generate text or image outputs. Generative vision-language models can be further divided into image generation models (e.g., diffusion models) \cite{ho2020denoising} and text generation models (e.g., MLLMs) \cite{wu2023multimodal}. Diffusion models, for instance, utilize U-Net \cite{ronneberger2015unet} and transformer architectures \cite{vaswani2023attention} for text-based image generation. MLLMs have emerged following the introduction of large language models (LLMs) \cite{brown2020language}. With the success of LLMs, researchers have explored the possibility of incorporating visual information as tokens into LLM models to leverage the previously learned knowledge in LLMs. Influential works in this area include LLaVA \cite{liu2024improved} and miniGPT-4 \cite{zhu2023minigpt4}.

In this work, we will focus on the data processing techniques used in diffusion models and MLLMs, because these models represent the forefront of multimodal machine learning, with significant advancements and wide-ranging applications. While our mainly focus on these recent models, we also cover some early discriminative models to provide a comprehensive overview of data processing techniques in the field.

\subsubsection{Multimodal Dataset}
Training datasets play a critical role in both model pretraining and finetuning. A typical dataset for vision-language models consists of pairs of images and text. The text and image pairs can have various relationships, corresponding to different tasks such as visual question answering (VQA) \cite{Zou_2020}, image retrieval \cite{liu2021image}, and text-to-image generation \cite{ho2020denoising}. The most commonly used dataset for pretraining vision-language models is the image captioning dataset, where the text is a description of the image. Examples include Conceptual Captions \cite{sharma-etal-2018-conceptual} and WIT \cite{Srinivasan_2021}.

Most of these data are collected from the internet but require significant processing. Recently, there has been a considerable effort in collecting and open-sourcing larger datasets, such as LAION-5B. The increase in both the scale and quality of training datasets has been observed to correlate with the improvement of vision-language models.

\section{Data Processing techniques}

\begin{figure*}[ht]
\begin{center}
\includegraphics[page=1, trim=2cm 4cm 2cm 4cm, clip, width=\linewidth]{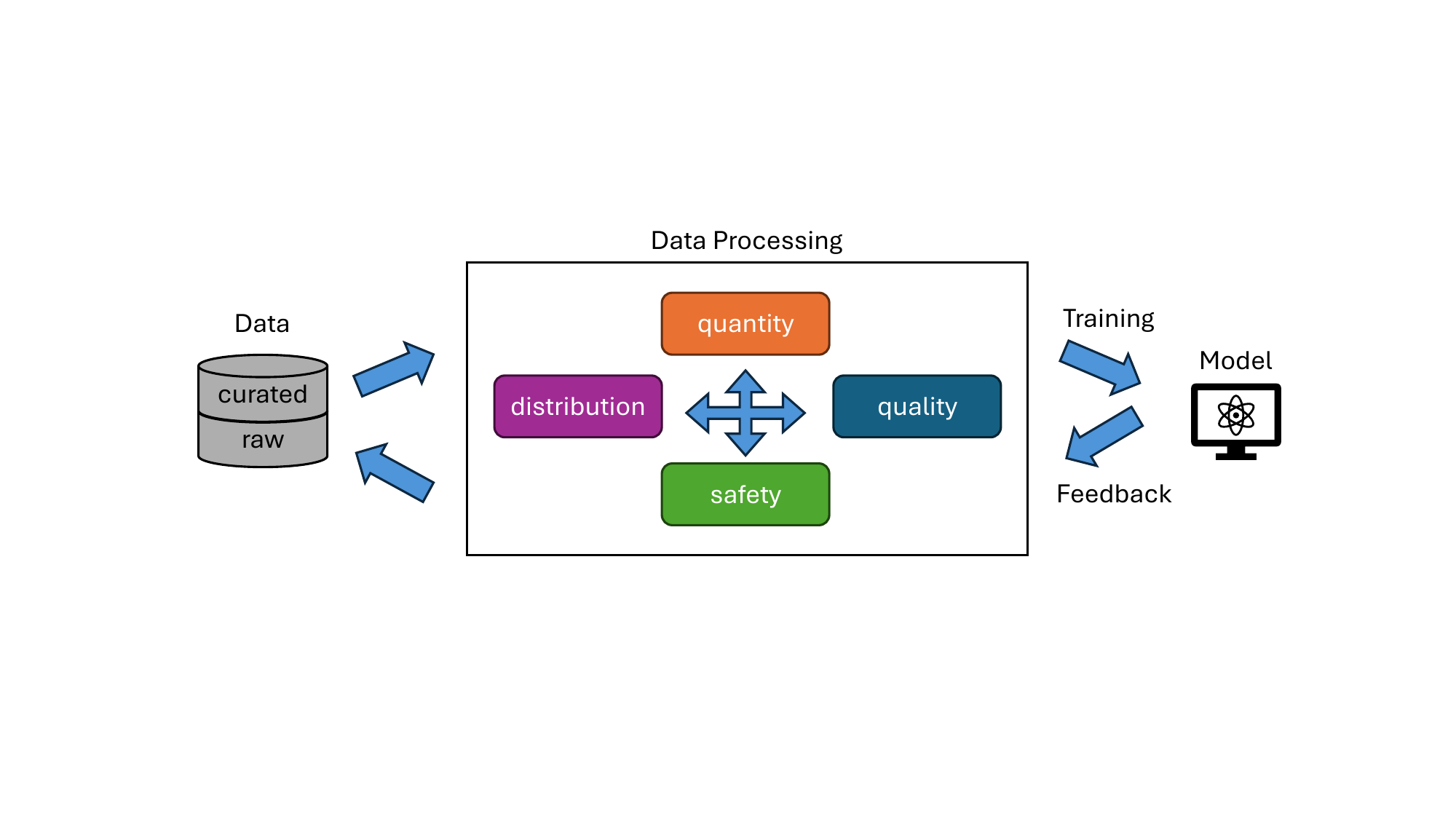}  
\caption{Data Processing Workflow}
\label{workflow}
\end{center}
\end{figure*}  

In this section, we discuss important data processing techniques used in multimodal generative models. To obtain a "good" training dataset, most efforts have focused on improving data distribution, data safety, data quality, and data quantity. We categorize the data processing techniques accordingly. In this work, we focus on three out of the four categories: data quality, data distribution, and data safety. Although data quantity is an important factor, it is not covered in this work as it relates to data sources, which are highly dependent on specific use cases.

We have outlined a standard workflow for data processing in training multimodal models in Figure \ref{workflow}. Generally, there are two types of data for used for training multimodal model: raw data and curated data. Raw data, such as LAION-5B, is collected directly from the internet and often comes in large quantities. However, it requires extensive cleaning to ensure high quality, safety, and usefulness for model training. On the other hand, curated datasets, such as MMC4 \cite{zhu2023multimodal}, are human-curated and require less effort in data cleaning, such as removing unsafe content. Nevertheless, it can still be beneficial to further improve data quality or optimize data distribution for the downstream task.

Once data is collected, we move to the data processing stage, where various techniques are used to enhance data quality, ensure data safety, and optimize or balance data distribution. The processed data is then used for model training. It is important to note that this is an iterative process. After the model is trained, we evaluate its performance and analyze failure cases. If certain issues are identified as caused by flaws in the training data, we return to the data processing step to adjust our methods. Additionally, if the data becomes insufficient at any point, we return to the initial stage to find more data sources and collect additional data.

In the following sections, we present a comprehensive collection of popular methods for quality filtering, distribution balancing, and data safety.

\subsection{Data Quality}
Data quality has always been a crucial factor in model training\cite{gunasekar2023textbooks}. Significant efforts have been made to improve the quality of single-modal data, such as WebText \cite{Radford2019LanguageMA}. In the multimodal setting, it is essential not only to enhance the quality of each individual modality but also to improve the alignment between image and text pairs. Most techniques for improving training data quality fall into two categories: data filtering and data augmentation.

\subsubsection{Filtering}
Quality filtering focuses on individual data samples, applying criteria to filter based on specific attributes. This can be done for text and images data, either separately or together.

\textbf{Image Filter} Image filters are most commonly used in diffusion based model training.

\begin{enumerate}
    \item Image property filter:
    Image resolution is a crucial aspect of quality; images with lower resolution generally exhibit lower quality and less information. A resolution filter is widely used in both MLLMs and diffusion-based models. Notable works like Stable Diffusion and EMU \cite{rombach2022high, podell2023sdxl} used resolution filters to process training datasets, retaining only images above a certain resolution threshold. Similarly, BLIP \cite{li2022blip} applies resolution filters to exclude low-quality images.
    
    Aspect ratio is another important factor. Images with extreme aspect ratios are difficult to process into the square shape which are typically required by training functions. Deepfloyd\cite{deep-floyd:IF} and EMU\cite{rombach2022high} filter their training images based on aspect ratio. SDXL \cite{podell2023sdxl} also emphasizes the importance of aspect ratio for improving model performance. 
    
    \item Advanced filter:  
    In addition to filters on the intrinsic properties on an image, filters can be defined based on abstract and advanced concepts. 
    
    Annotator Guided Filter: Some models use expert annotators to filter data based on aesthetic quality. EMU, for instance, relies on annotators with photography expertise to select the highest quality images, which are then used to fine-tune the model. This results in generated images that are significantly preferred for their visual appeal and text faithfulness.
    
    Model Guided Filter: Other works, like Stable Diffusion1.5 \cite{rombach2022high} and Stable Diffusion2.1\cite{sd21}, use smaller models to score each image and filter them based on these scores. These scores can represent aesthetic quality or watermark intensity.
    
    \item Source filter: The data source can indicate the quality of the data. For example, images from social platforms (e.g., Instagram) often come with user feedback such as likes or reposts, which can indicate the quality of the image from the user's perspective. EMU uses the number of likes on an image as a filter to select high-quality images.   
\end{enumerate}

\textbf{Text Filter} Text filtering is more common in MLLM-related research.
\begin{enumerate}
    \item Annotator Guided Filter: Annotator guided filters are also common in text filtering. In miniGPT-4 \cite{zhu2023minigpt4}, training data for the second stage is filtered and refined by humans with assistance from ChatGPT.

    \item Model Guided Filter: In miniGPT-4, ChatGPT is used as an initial filter to detect errors in image captions. The authors define frequent errors in the captions and instruct ChatGPT to identify such errors in the dataset. In BLIP \cite{li2022blip}, a model is trained specifically to filter out captions that are not well-aligned with the given image.
    
    \item Image-Text Alignment Filter: Multimodal alignment filtering is also important. CLIP \cite{radford2021learning} models are widely used to measure alignment across different modalities, especially for image and text. In Stable Diffusion, SDXL, and EMU, image-text pairs are filtered based on their CLIP scores, improving the model's capability of following user instructions.
\end{enumerate}

\begin{table*}[htbp]
\caption{Data Processing Techniques and Datasets in Different Models}
\centering
\begin{tabular}{|p{1.9cm}|p{2cm}|p{2.5cm}|p{2.5cm}|p{2.5cm}|p{2cm}|p{1.8cm}|}
\hline
\textbf{Model} & \textbf{Model Category} & \textbf{Safety} & \textbf{Quality} & \textbf{Balance} & \textbf{Pretraining Dataset} & \textbf{Finetuning Dataset} \\
\hline
LLaVA & MLLM & - & - & Noun Phrase Popularity, use ChatGPT to generate Instruct-158K dataset for finetuning & Conceptual Captions & Instruct-158K, ScienceQA \cite{lu2022learn} \\
\hline
LLaVA-1.5 & MLLM & - & - &  Unknown & OKVQA\cite{marino2019okvqa}, A-OKVQA\cite{schwenk2022aokvqa} , OCRVQA\cite{mishraICDAR19}, TextCaps\cite{sidorov2020textcaps}, Visual Genome, RefCOCO\cite{kazemzadeh-etal-2014-referitgame}, ShareGPT & Downstream tasks \\
\hline
MiniGPT4 & MLLM & - & Use early model to generate data, then use ChatGPT and human filtering to get 3,500 image-text pairs & - & Conceptual Captions, SBU\cite{sbu}, LAION & 3.5k image-text pairs \\
\hline
CLIP & Dual Encoder & Bias toward age, gender, and race is reported, no mitigation is mentioned & Unknown & Noun Phrase Popularity & Closed Source Dataset & - \\
\hline
BLIP & MLLM & N/A & Model guided filtering and generation to get CapFilt dataset & - & COCO, Visual Genome\cite{krishna2016visual}, Conceptual Captions, Conceptual 12M \cite{changpinyo2021conceptual}, SBU & CapFilt dataset \\
\hline
BLIP-2 & MLLM & N/A & Model guided filtering and generation to get CapFilt, use CLIP to filter image-text aligned data & - & COCO, Visual Genome\cite{krishna2016visual}, Conceptual Captions, Conceptual 12M \cite{changpinyo2021conceptual}, SBU, LAION-400M & Downstream tasks \\
\hline
Stable Diffusion 1.5 & Diffusion Models & NSFW Filter, Watermark Filter, Image-Text Alignment & Resolution, Aesthetic & - & LAION-5B & - \\
\hline
Stable Diffusion 2.1 & Diffusion Models & NSFW Filter, Watermark Filter & Resolution, Aesthetic, Image-Text Alignment & - & LAION-5B & - \\
\hline
Stable Diffusion XL & Diffusion Models & NSFW Filter, Watermark Filter, Image-Text Alignment & Unknown & Unknown & Private Dataset & - \\
\hline
EMU & Diffusion Models & NSFW Filter & Resolution, Aesthetic, Aspect Ratio, Source Filter & Image classification & Private Dataset & Closed Source Dataset \\
\hline
Deep Floyd & Diffusion Models & NSFW Filter & Resolution, Aesthetic, Aspect Ratio & - & LAION-5B & - \\
\hline
\end{tabular}

\label{tab:model-comparison}

\end{table*}

\subsubsection{Augmentation}

With the advancement of vision-language models, it has become feasible to generate high-quality text and image data using early model versions or closed-source APIs (ChatGPT). These generated data can be used to further train and improve models.

Many works in this field focuses on generating high-quality text captions for images. For instance, BLIP \cite{li2022blip} uses captioning models initialized from early BLIP model checkpoints to generate captions for web images. Similarly, MiniGPT-4 \cite{zhu2023minigpt4} employs early checkpoints to generate descriptions for finetuning datasets, with ChatGPT-4 \cite{openai2024gpt4} further refining these captions. In the training of LLaVA-1.5 \cite{liu2024improved}, large-scale GPT-annotated datasets such as LAION-GPT-V and ShareGPT-4V \cite{chen2023sharegpt4v} are utilized to enhance performance.

\subsection{Data Distribution}
Distribution balancing focuses on producing a well-balanced and diverse dataset based on predefined criteria. Data sampling is widely used in multimodal data processing. While typically used for filtering higher quality data, it is also useful for balancing the training data distribution. Text and images are the two primary data sources for vision-language models, corresponding to two dimensions for filtering and balancing.

\subsubsection{Image Oriented Balancing} 

In the EMU model \cite{dai2023emu}, images are balanced by leveraging visual concepts from different domains using an image classification model \cite{yalniz2019billionscale}. Image deduplication is another common practice in data processing. In the MMC4 \cite{zhu2023multimodal} dataset, models like phash \cite{Zauner2010ImplementationAB}, which identify visually similar images, are used to remove duplicates. The MMC4 dataset is popular for developing vision-language models such as OpenFlamingo \cite{awadalla2023openflamingo}. 

\subsubsection{Text Oriented Balancing}

Text-oriented data balancing is more common in MLLMs. In LLaVA, image captions are sampled based on noun phrase frequency. The goal is to exclude rare noun phrases and reduce overly common ones. For a text caption to be included, the frequency must be at least three. Meanwhile, if a noun phrase appears too frequently (more than 100 times), only 100 captions containing that phrase are randomly selected. In addition to sampling, LLaVA model uses GPT-4 and ChatGPT to generate in-depth and diversity conversations from COCO images\cite{lin2015microsoft}. A similar approach has been found in creating pretraining dataset for CLIP\cite{radford2021learning} model. To ensure a coverage of a broad set of visual concepts, the authors use frequently occurring words from Wikipedia to construct base queries. These queries filter text captions, ensuring that only those including one of the query words are included in the dataset.

\subsection{Data Safety}
Data safety is critical in developing ethical and trustworthy models. In this section, we define two problems of data safety: data toxicity and social bias and unfairness. While there is no consensus definition of toxic data, it generally includes violent, pornographic, offensive, and unethical content, also known as Not Safe For Work (NSFW). Biased and unfair data are more subtle and refer to social biases or stereotypes in the data. The definition of bias and fairness is more ambiguous than NSFW data, as it often depends on specific downstream tasks and cultural background.

\subsubsection{Toxic Data}

For multimodal datasets, toxic data can appear in both image and text content. Wordlists such as LDNOOBW \cite{LDNOOBW} are created to filter harmful text. For image safety, CLIP embeddings are often used to compare images with a list of toxic text, and images with high similarity scores are tagged as unsafe. For instance, in the LAION-5B dataset, the authors use both the Q16 classifier \cite{schramowski2022machines} and their own specialized pornographic and sexual content classifier, both based on CLIP embeddings, to filter inappropriate images. Stable Diffusion 1.5 and 2.1 use this LAION dataset with the NSFW filter.

\subsubsection{Social Bias and Fairness}

Addressing bias and fairness in MLLMs is an emerging research area. a number of work has been done to analyze and evaluate social bias in language models \cite{liang2021understanding, gallegos2024bias, 10.1162/coli_a_00524}, and similar techniques can be applied to processing text in multimodal training data. Counterfactual data augmentation (CAD) is a common technique used to balance representation bias in text data. For example, in \cite{lu-etal-2021-neurologic}, CAD is used to balance bias by flipping gendered pronouns in existing data to generate new samples. In diffusion models, social bias and stereotypes have been recognized as problems \cite{luccioni2023stable, bansal2022texttoimage, naik2023social}. However, most debiasing work is focused on model training rather than data processing \cite{berg2022prompt}.

To summarize, while eliminating harmful content for data safety has been largely successful, more work is needed to mitigate social bias and promote fairness in multimodal models.

\section{Comparison of Data Processing Techniques in Different Models}

Table \ref{tab:model-comparison} provides a summary of the MLLMs and diffusion models discussed in the previous section. It provides a clear comparison of the techniques used in different types of models.

According to Table \ref{tab:model-comparison}, diffusion models focus more on image data quality, such as aesthetics and resolution, while MLLMs focus more on text data quality and its alignment with images. Diffusion models often require large scale data such as LAION-5B, where data safety filters are applied to remove toxic data. MLLMs frequently use curated datasets which is smaller and require fewer additional safety handling in data processing. This is because the LLM model inside MLLMs have already undergo training with large scale text data. So, a majority of the work in MLLM is only to properly project image into text tokens. Both diffusion models and MLLMs leverages distribution balancing techniques to obtain a representative and diverse dataset in terms of text or image topics.

Data augmentation has become a popular technique for generating high-quality data for model training, especially for MLLMs. Since MLLMs leverage existing LLM frameworks such as LLaMA \cite{touvron2023llama}, which is equipped substantial capabilities in understanding and generating text, these models do not require a highly diverse set of text. However, for the model to understand the association between text and image, it is crucial that the text is well-aligned with the image. Text alignment is a core aspect of data quality in MLLM training. It has been shown, in the training of models like MiniGPT-4 and LLaVA-next, that synthetic data generated from LLMs or early checkpoints results in better text alignment than unprocessed raw data. For diffusion models, using synthetic image-text data for training is less common.

In the data processing techniques used by both diffusion models and MLLMs, there is a trend toward increased use of model-based filters, such as using ChatGPT or NSFW classifiers to refine training data. However, human annotators are still used as the final judges of data quality. Especially during the finetuning stage, which requires extremely high-quality data, the inclusion of human expert annotators can still lead to significant improvements in data quality.

\section{Conclusion}
In this paper, we discussed common data processing techniques for multimodal models, specifically vision-language models. We proposed a framework that categorizes these techniques from four perspectives: quantity, quality, safety, and distribution. We compared the techniques used in image generation models—diffusion models—and text generation models—MLLMs.

Data processing is an iterative process. The choice of processing methods depends on specific tasks as well as model performance, and there are no standard steps to follow for training data processing. While this survey does not provide an exhaustive list of all available data processing techniques, we have carefully selected the techniques used by the most influential models. We hope this work provides a useful guideline for deciding on methods for data processing.

\bibliographystyle{IEEEtran}
\bibliography{sample}

\end{document}